\documentclass[10pt,twocolumn,letterpaper]{article}

\usepackage[pagenumbers]{cvpr}

\definecolor{cvprblue}{rgb}{0.21,0.49,0.74}
\usepackage[pagebackref,breaklinks,colorlinks,allcolors=cvprblue]{hyperref}

\definecolor{myblue}{rgb}{.39,.58,.93}
\definecolor{runpei_orange}{HTML}{F35F27}

\def\Ours{MAGI\xspace}

\usepackage{multirow}
\usepackage{pifont}
\newcommand{\cmark}{\ding{51}} 
\newcommand{\xmark}{\ding{55}}

\title{Taming Teacher Forcing for Masked Autoregressive Video Generation}

\author{
  Deyu Zhou\textsuperscript{$1$}\quad
  Quan Sun\textsuperscript{$2$}\quad
  Yuang Peng\textsuperscript{$2,4$}\quad
  Kun Yan\textsuperscript{$2$}\quad
  Runpei Dong\textsuperscript{$3$}\quad 
  Duomin Wang\textsuperscript{$2$}\\
  Zheng Ge\textsuperscript{$2$}\quad
  Nan Duan\textsuperscript{$2$}\quad
  Xiangyu Zhang\textsuperscript{$2$}\quad
  Lionel M. Ni\textsuperscript{$1$}\quad
  Heung-Yeung Shum\textsuperscript{$5$}
  \\[0.08in]
  \textsuperscript{$1$}HKUST(GZ) \qquad
  \textsuperscript{$2$}StepFun \qquad
  \textsuperscript{$3$}UIUC \qquad
  \textsuperscript{$4$}THU \qquad
  \textsuperscript{$5$}HKUST \\
  {\scshape
\textbf{\href{https://magivideogen.github.io/}{MAGI-Video-Generation.Github.Io}}
}
}

\begin{document}
\maketitle

\begin{abstract}
We introduce MAGI, a hybrid video generation framework that combines masked modeling for intra-frame generation with causal modeling for next-frame generation. Our key innovation, Complete Teacher Forcing (CTF), conditions masked frames on complete observation frames rather than masked ones (namely Masked Teacher Forcing, MTF), enabling a smooth transition from token-level (patch-level) to frame-level autoregressive generation. CTF significantly outperforms MTF, achieving a \textbf{+23\%} improvement in FVD scores on first-frame conditioned video prediction. To address issues like exposure bias, we employ targeted training strategies, setting a new benchmark in autoregressive video generation. Experiments show that MAGI can generate long, coherent video sequences exceeding 100 frames, even when trained on as few as 16 frames, highlighting its potential for scalable, high-quality video generation.
\end{abstract}





\begin{table*}[ht]
    \small
    \centering
    \caption{\textbf{Comparison of autoregressive paradigms across different methods.} We compare MAGI with various methods based on temporal attention mechanisms (how each frame attends to others), support for KV Cache~\cite{KeepCostDown24} and variable context lengths, observation completeness (the completeness of the context frames used for conditioning), and the prediction granularity (patch or frame).
    $^\dagger$: MAGVIT predicts masked tokens of multiple frames simultaneously. Genie~\cite{bruce2024genie} and MAGI predict multiple tokens of single frame simultaneously. 
    }
    \label{tab:comparison_method}
    \resizebox{\linewidth}{!}{
    \begin{tabular}{l l c c c c}
    \toprule[0.95pt]
    \textbf{Prediction Granularity} & \textbf{Method} & \textbf{Temporal Attention} & \textbf{KV Cache} & \textbf{Var. Context} & \textbf{Observation} \\
    \midrule[0.6pt]

    \textbf{Patch} & VideoGPT~\cite{yan2021videogpt}, Phenaki~\cite{villegas2022phenaki}, Omni~\cite{wang2024omnitokenizer} & Causal & \cmark & \cmark & Complete \\

    \midrule[0.6pt]

    \multirow{4}{*}{\textbf{Frame}} & MAGVIT$^\dagger$~\cite{yu2023magvit}, GameNGen~\cite{valevski2024diffusion} & Bidirectional & \xmark & \xmark & Complete \\

    & Diffusion Forcing~\cite{chen2024diffusion} & Causal & \cmark & \cmark & Noisy \\

    & Genie~\cite{bruce2024genie} & Causal & \cmark & \cmark & Masked \\

    & \textbf{MAGI (with CTF)} & Causal & \cmark & \cmark & Complete \\

    \bottomrule[0.95pt]
    \end{tabular}
    }
    \vspace{-8pt}
    \end{table*}
\section{Introduction} \label{sec:intro}
Generating \textit{order} matters in autoregressive image generation. While existing approaches mostly reply on simple raster-scan order, recent studies~\cite{li2024autoregressive, fan2024fluid} have demonstrated superior results through alternative generation strategies. 
These include randomized spatial order with bi-directional attention~\cite{li2024autoregressive,chang2022maskgit,yu2023magvit} and generation along increased scales~\cite{fan2024fluid,chang2023musetexttoimagegenerationmasked}. 
Despite its fundamental importance, in the realm of autoregressive video generation, however, the discussion of generation order has been largely overlooked.

Existing autoregressive video generation methods can be categorized into two groups according to the prediction granularity, as shown in \cref{tab:comparison_method}.
The first group, exemplified by MAGViT~\cite{yu2023magvit}, adopts a masked modeling approach\footnote{We follow the definition in MAR~\cite{li2024autoregressive}, which considers masked image generation as a form of auto-regressive generation with a certain order.} where visual tokens are generated in descending order according to their logit probabilities. 
While this approach uses bi-directional attention across frames, it presents two significant limitations: substantial computational overhead during inference since it is unable to utilize KV Cache~\cite{AdaptiveKVCache24,KeepCostDown24}, and neglect temporal causality between consecutive frames. 
These oversights are particularly noteworthy given the remarkable success of causal modeling in large language models.
Some other methods in this group~\cite{bruce2024genie,chen2024diffusion} have explored frame-level prediction with casual temporal attention.
However, these methods are limited due to incomplete history observation, which causes an intrinsic training-inference gap.

The second group comprises fully autoregressive approaches operating on visual patches~\cite{yan2021videogpt,wu2022nuwa,kondratyuk2023videopoet,wang2024emu3}, exemplified by recent works like VideoPoet~\cite{kondratyuk2023videopoet} and Emu3~\cite{wang2024emu3}.
Despite their successful temporal modeling, these methods have not incorporated contemporary advances from image generation research in intra-frame generation. Instead, they continue to rely on raster-scan order - an approach demonstrated sub-optimal in image generation~\cite{li2024autoregressive}.

\begin{figure*}[t!]
    \centering  
    \includegraphics[width=0.8\linewidth]{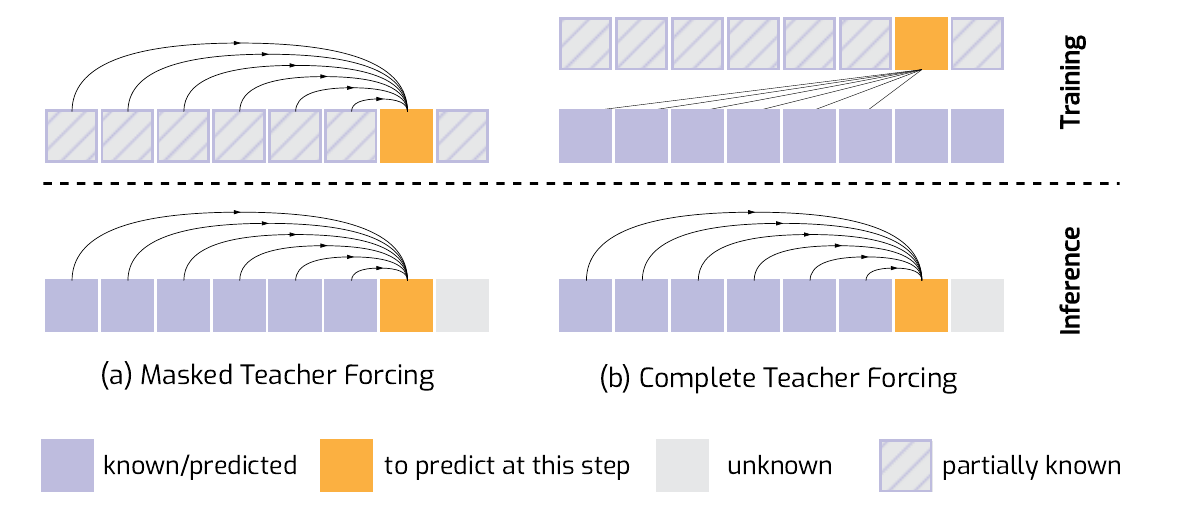}

    \vspace{-5pt}
    \caption{\textbf{Conceptual comparison of \textit{masked teacher forcing} and our proposed \textit{complete teacher forcing} mechanisms in auto-regressive video generation.}
    The blocks here illustrate individual frames.
    These two mechanisms differ in how history observation frames are used to condition next frame generation during training.
    \textbf{(a):} The observation frames are \textbf{\textit{masked frames}}~\cite{MAE22,bruce2024genie} that are \textit{partially} observable. 
    \textbf{(b):} Every frame is generated conditioned on \textit{fully} observable history frames.
    }
    \label{fig:onecol}
\end{figure*}

In this paper, we introduce \textbf{M}asked \textbf{A}utoregressive video \textbf{G}enerat\textbf{I}on (\textbf{MAGI}), a hybrid framework that synergizes the strengths of both video generation paradigms by combining masked modeling for intra-frame generation with causal modeling for inter-frame dependencies. We start with a naive implementation. This initial implementation explores a straightforward approach where tokens in each next frame are generated conditioned on previous masked frames (Figure~\ref{fig:onecol}(a)). While this can be viewed as a natural extension of MaskGIT~\cite{chang2022maskgit} and Muse~\cite{chang2023musetexttoimagegenerationmasked} to autoregressive video generation, our analysis reveals a fundamental limitation in its implementation of \textit{teacher forcing} (TF)~\cite{williams1989} - a crucial training technique in autoregressive models for avoiding exposure bias and improving scalability. 
To be specific, TF replaces predicted tokens with ground-truth tokens during training. 
However, our initial approach, which we term as \textbf{Masked Teacher Forcing} (MTF), deviates from this principle by substituting ground-truth tokens with mask tokens. 
Despite similar approaches being explored in interactive generative models (\eg, Genie~\cite{bruce2024genie}), we identify a critical training-inference discrepancy: during training, next-frame generation is conditioned on a mixture of visible and masked tokens, while during inference, it relies solely on complete visual tokens from previous frames. Our experiments demonstrate that this inconsistency compromises generation quality, especially motion coherency. 

To address these limitations, we propose \textbf{Complete Teacher Forcing} (CTF), an advanced paradigm that maintains training-test consistency by conditioning next-frame generation on complete visual observations from history frames during training. This is achieved through a novel mechanism that prepends complete video tokens to the masked input sequence and employs a carefully designed attention mask. Empirical results show that CTF better captures motion, outperforming MTF by \textbf{23\%} in terms of FVD scores on first-frame conditioned video prediction.

In addition, we tackle the persistent challenge of exposure bias~\cite{CloseLookAtExposureBias19} in autoregressive video generation through two complementary strategies: dynamic interval training and noise injection~\cite{valevski2024diffusion}. 
Dynamic interval training refers to randomly sampling frames with varying intervals during training, which introduces diversity into the data distribution and helps the model to better generalize to different temporal frequencies. Dynamic noise injection, on the other hand, involves adding random noise to observation frames during training, which helps improve the model's robustness by simulating the possible errors may occur in inference. Combined with these two techniques, \Ours establishes a robust baseline for autoregressive video generation. Through extensive evaluations, we demonstrate that \Ours achieves superior video generation quality and length scalability.
For example, \Ours can generate coherent video sequences that exceed 100 frames, even when trained on sequences as short as 16 frames. 
This underscores the potential of our approach for scalable video generation. 
\section{Background and Problem Statement}
\label{sec:background}
\subsection{Autoregressive Video Generation}
Given a video $V \in \mathbb{R}^{N\times H \times W \times 3}$, 
the sequence of $N$ frames with height $H$ and width $W$ can be denoted as $V =\{f_i\in\mathbb{R}^{H\times W\times 3}|i=1,\dots,N\}$.
Each frame can be further divided into $m$ patches, represented as $\{\mathbf{x}_{11}, \mathbf{x}_{12}, \ldots, \mathbf{x}_{1m}, \ldots, \mathbf{x}_{n1}, \mathbf{x}_{n2}, \ldots, \mathbf{x}_{nm}\}$. Autoregressive video generation models primarily differ in the granularity of causality (patch-level vs. frame-level) and the training paradigm (teacher forcing), discussed below.  

\noindent \textbf{Patch-level Methods}  
Patch-level autoregressive models~\cite{yan2021videogpt, wang2024emu3} operate by modeling the video on a patch-by-patch basis. The likelihood of generating entire video with a model parameterized by $\theta$ is decomposed into the product of posterior of each patch conditioned all preceding patches:  
\begin{align}
p(V) = \prod_{t=1}^{N \times m} p(\mathbf{x}_t \mid \mathbf{x}_1, \mathbf{x}_2, \ldots, \mathbf{x}_{t-1}; \theta),
\end{align}
where $p_t$ represents a patch in a predefined ordering. 
The model generates each patch sequentially by conditioning on all previously generated patches, capturing both spatial and temporal dependencies. 

\noindent \textbf{Frame-level Methods} 
Frame-level autoregressive models~\cite{bruce2024genie, valevski2024diffusion} model the video at the granularity of entire frames. 
Similarly, the likelihood of video modeling is:  
\begin{align}
p(V) = \prod_{t=1}^{n} p(f_t \mid f_1, f_2, \ldots, f_{t-1}; \theta),
\end{align}
where \(f_t\) is the entire frame generated conditioned on all history frames. 
In this fashion, the patches in each frame are generated in parallel.

\noindent \textbf{Teacher Forcing}
Teacher forcing~\cite{williams1989} is a widely adopted training strategy in autoregressive models, where the model conditions on the \textit{true} previous frame or patch rather than its own predictions~\cite{williams1989,ProfessorForcing16}. 
This method is commonly used in patch-level autoregressive video models~\cite{yan2021videogpt,wu2022nuwa,kondratyuk2023videopoet, wang2024omnitokenizer} as well as in language models~\cite{radford2018improving,radford2019language}, where the model is supervised through next token or patch prediction, effectively ``shifting'' one token or patch at a time during training.

\subsection{Can we achieve frame-level teacher forcing?}
While teacher forcing is effective in autoregressive language models~\cite{radford2018improving,radford2019language}, its application to frame-level video generation is largely unexplored. Few works adopt a straightforward “shift one frame” paradigm for frame-level video models.

Instead, existing methods often rely on masked frame prediction, where masked frames are conditioned on other frames. For example, Genie~\cite{bruce2024genie} extends MaskGIT~\cite{chang2022maskgit} to video generation by conditioning masked frames on other masked teacher frames—a strategy we refer to as \textit{Masked Teacher Forcing} (MTF). Similarly, Diffusion Forcing~\cite{chen2024diffusion} introduces temporal causal attention with a novel noise strategy but conditions each frame on \emph{noisy} teacher frames, further diverging from traditional teacher forcing. GameNGen~\cite{valevski2024diffusion} employs bidirectional diffusion models with binary mask tokens to represent masked and conditional frames; however, it relies on fixed-length conditional frames, limiting autoregressive flexibility and preventing the use of efficient mechanisms like KV Cache—one notable advantage of traditional autoregressive methods.
Generating videos autoregressively at the \emph{patch level} is also suboptimal because inter-frames lack clear raster-scan spatial causality. In contrast, videos exhibit strong temporal causality, suggesting that predicting videos frame by frame, while generating the tokens within each frame in parallel, is more effective.

To address these issues, \Ours fully leverages teacher forcing for frame-level autoregressive video generation, enabling more flexible, efficient, and scalable video generation.

\section{Approach}
\label{sec:method}
Existing frame-level autoregressive video generation models, such as GameNGen~\cite{valevski2024diffusion}, typically use binary masks to distinguish between frames used for prediction and those used for conditioning. These models are trained to predict future frames conditioned on a fixed-length sequence of preceding frames. This fixed conditioning contrasts with the greater flexibility of autoregressive language models, which condition on prefixes of variable length. In this work, we leverage MAR~\cite{li2024autoregressive}, a state-of-the-art masked image generation framework, and investigate crucial design choices for masked autoregressive video generation. Our focus includes exploring different autoregressive paradigms and methods for mitigating exposure bias and error accumulation during autoregressive inference.
\paragraph{Masked Teacher Forcing \textit{vs.} Complete Teacher Forcing.}
A natural extension of masked image generation~\cite{chang2022maskgit, li2024autoregressive} to autoregressive video prediction is to modify the attention mechanism~\cite{bruce2024genie}, incorporating causal temporal attention during the training phase for masked frames. 
We refer to this approach as Masked Teacher Forcing (MTF), as illustrated on the right of Fig.~\ref{fig:onecol}. However, this method introduces a significant training-inference gap: during training, the model attends to a high mask ratio, while at inference time, it must predict future frames conditioned on previously generated unmasked ones. This mismatch between training and inference undermines the model's ability to generate realistic videos, ultimately limiting its performance.
\begin{figure}[t!]
  \begin{center}
  \includegraphics[width=\linewidth]{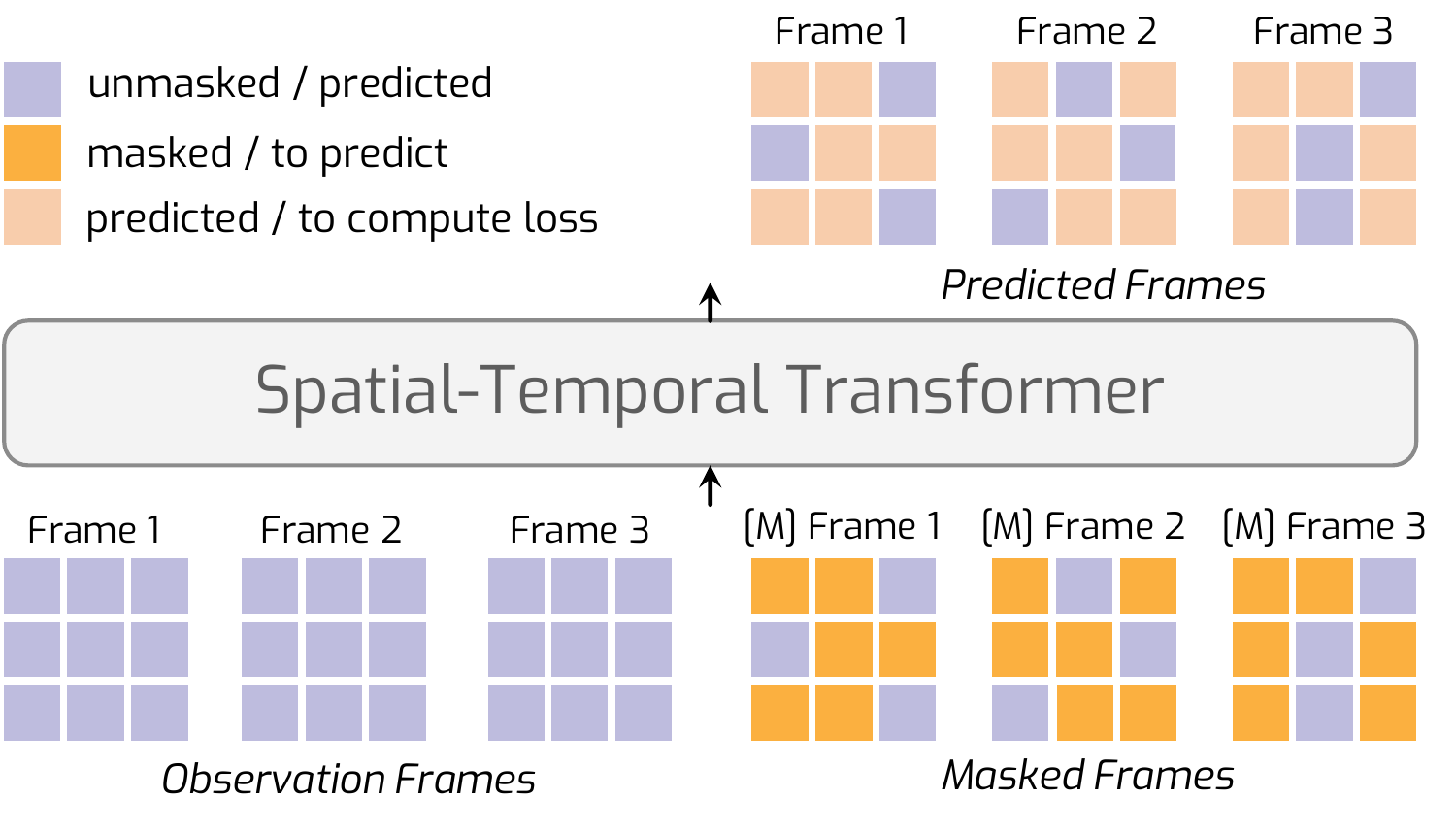}
  \vspace{-20pt}
  \caption{\textbf{Overview of MAGI video generation framework}. MAGI receives observation frames and corresponding masked frames as inputs, enabling autoregressive video generation with Complete Teacher Forcing (\cref{sec:method}).  
  } \label{fig:framework}
  \end{center}
  \vspace{-15pt}
\end{figure}

The formal formulation of MTF is as follows:  
\begin{align}
p(f^m_j \mid f^m_1, f^m_2, \ldots, f^m_{j-1}; \theta), & \quad j \in \{1, 2, \dots, n\}.
\end{align}
where \(f^m_j\) represents the \(j\)-th masked frame, while the frames \(f^m_1, f^m_2, \ldots, f^m_{j-1}\) are the previously predicted masked frames. In this formulation, during training, each masked frame attends to previously masked frames and itself. However, the approach introduces a significant gap between the training and the inference phase, because the model must predict masked frames conditioned on previously unmasked generated ones at the inference stage.

To address this issue and more effectively bridge the gap between training and inference, we propose a novel autoregressive paradigm, termed \textbf{Complete Teacher Forcing} (CTF), as shown on the left of~\cref{fig:onecol}. Unlike MTF, CTF conditions on unmasked observed frames during training, predicting each masked frame.  Formally, let the observation frames be denoted as \(\{f_1, f_2, \ldots, f_n\}\), where \(n\) is the number of frames. The autoregressive formulation for predicting the \(j\)-th masked frame, \(f^m_j\), is conditioned on both the previous unmasked observation frames and itself, \ie,
\begin{align}
p(f^m_j \mid f_1, f_2, \ldots, f_{j-1}; \theta), \quad j \in \{1, 2, \dots, n\},
\end{align}
where \(p(f^m_j \mid \cdot; \theta)\) is the autoregressive model. 
Note that \(f^m_1\) is conditioned only on itself, which resembles image generation. For all \(j > 1\), \(f^m_j\) attends to both the previous unmasked observation frames and itself, making this approach consistent between training and inference.

\paragraph{Addressing Exposure Bias \& Error Accumulation}
Although CTF reduces the gap between training and inference, it still faces challenges related to \textit{exposure bias}~\cite{zhang2019bridging} and \textit{error accumulation}~\cite{laskey2017dart}. Specifically, the model may struggle to generate realistic video sequences during inference when relying on self-predicted frames.
\begin{figure}[t!]
    \centering  
    \includegraphics[width=0.618\linewidth]{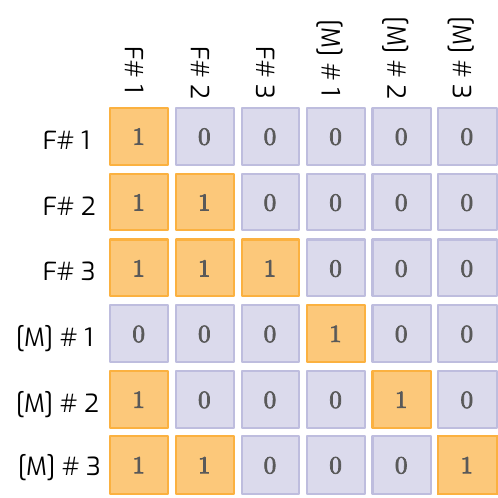}
    \caption{\textbf{Temporal attention mask} in our CTF during training. 
    The attention within observation frames \textit{causal}, while the attention within masked frames are \textit{atrous}.
    In this fashion, each masked frame attends to itself and unmasked history observation frames. During inference, a standard causal mask is employed, where each frame attends only to previously generated frames.
    }
    \label{fig:masking_strategy}
    \vspace{-10pt}
\end{figure}

To mitigate these issues, we introduce two strategies:
\begin{itemize}
    \item \textit{\textbf{Dynamic interval training.}} Video clips are sampled with random frame intervals, which forces the model to learn longer temporal dependencies and larger motion ranges, thus improving prediction stability for long videos.
However, we find that a vanilla usage of this strategy leads to videos generated with uncontrollable motion range during inference.
For example, prediction with a random motion range leads to unsatisfactory video generation results on UCF-101 that requires generation in 25 FPS.
To support controllable generation to handle varying user-specified frame intervals, we introduce \textit{learnable interval embeddings} which encodes different intervals into specific embeddings like positional encoding~\cite{vaswani2017attention}.
Specifically, the $x$-th interval embedding will be added to hidden states when sampling with an interval $x$ which enables the model's awareness of the desired generation interval.
    \item \textbf{\textit{Dynamic noise injection.}}
As pointed out by \citet{valevski2024diffusion}, there is a domain shift caused by teacher forcing and auto-regressive modeling, and thus data corruption using an injected noise during training is useful.
Inspired by this practice, we also adopt a dynamic noise injection strategy.
In addition, similar to interval embeddings, a learnable noise level embedding is concatenated with hidden states as the model inputs.
Both techniques are crucial for improving the robustness of CTF by ensuring better generalization and stability during inference.
\end{itemize}

\section{Architecture}
We introduce \Ours, as shown in \cref{fig:framework}, a novel family of masked autoregressive video generation models, which incorporates the techniques outlined above.
We implement \Ours with a Transformer architecture, detailed as follows.

\paragraph{Transformer Decoder \& Temporal Attention}
\Ours employs a stack of spatial-temporal Transformer blocks consisting of interleaved 2D spatial attention layers and 1D temporal attention layers.
As shown in~\cref{fig:framework}, we prepend the complete observation frames with masked frames as the input of our Transformer.
For CTF, we design a special temporal attention mask, as shown in ~\cref{fig:masking_strategy}. 
In the temporal layer, each frame only attends to itself and its preceding observation frames. The observation frames only attend to previous frames. Thus, the attention paradigm of CTF is consistent between training and inference.

\noindent \textbf{Diffusion Head~\cite{li2024autoregressive}}~~
Atop the Transformer decoder, we stack multi-layer perceptron (MLP) layers as the diffusion head, following MAR~\cite{li2024autoregressive}. 
This component predicts masked tokens through a denoising diffusion procedure~\cite{ho2020denoising}, enhancing the model's capacity for autoregressive generation.

\noindent \textbf{Learnable Positional Embeddings}~~
To distinguish between masked and unmasked frames, we introduce two distinct learnable positional embeddings: one for the masked frames and another for the observation frames. Additionally, we adopt a learnable positional embedding for the diffusion head, as proposed in MAR~\cite{li2024autoregressive}.

\noindent \textbf{Spatial and Temporal Positional Embeddings}~~
We utilize sinusoidal positional embeddings~\cite{vaswani2017attention} for both 2D spatial and 1D temporal encodings, ensuring that the model effectively captures spatiotemporal relationships across frames.

\noindent \textbf{Frame Interval Embeddings}~~
As part of our dynamic interval training strategy, we introduce a learnable interval embedding, with a vocabulary length of 25, which covers frame intervals ranging from 1 to 25. This embedding allows the model to capture long-range temporal dependencies across varying frame intervals.

\noindent \textbf{Noise Level Embeddings}~~
To support dynamic noise injection, we incorporate random Gaussian noise into the observation frames during training, following GameNGen~\cite{valevski2024diffusion}.
We use a noise level with a range of $[1,5]$, which is encoded through a learnable vocabulary embedding, with a dimension the same as the Transformer's hidden dimension.
This enables the model to adapt to different noise levels during the denoising process.

\begin{figure*}[t!]
    \centering  
    \includegraphics[width=\linewidth]{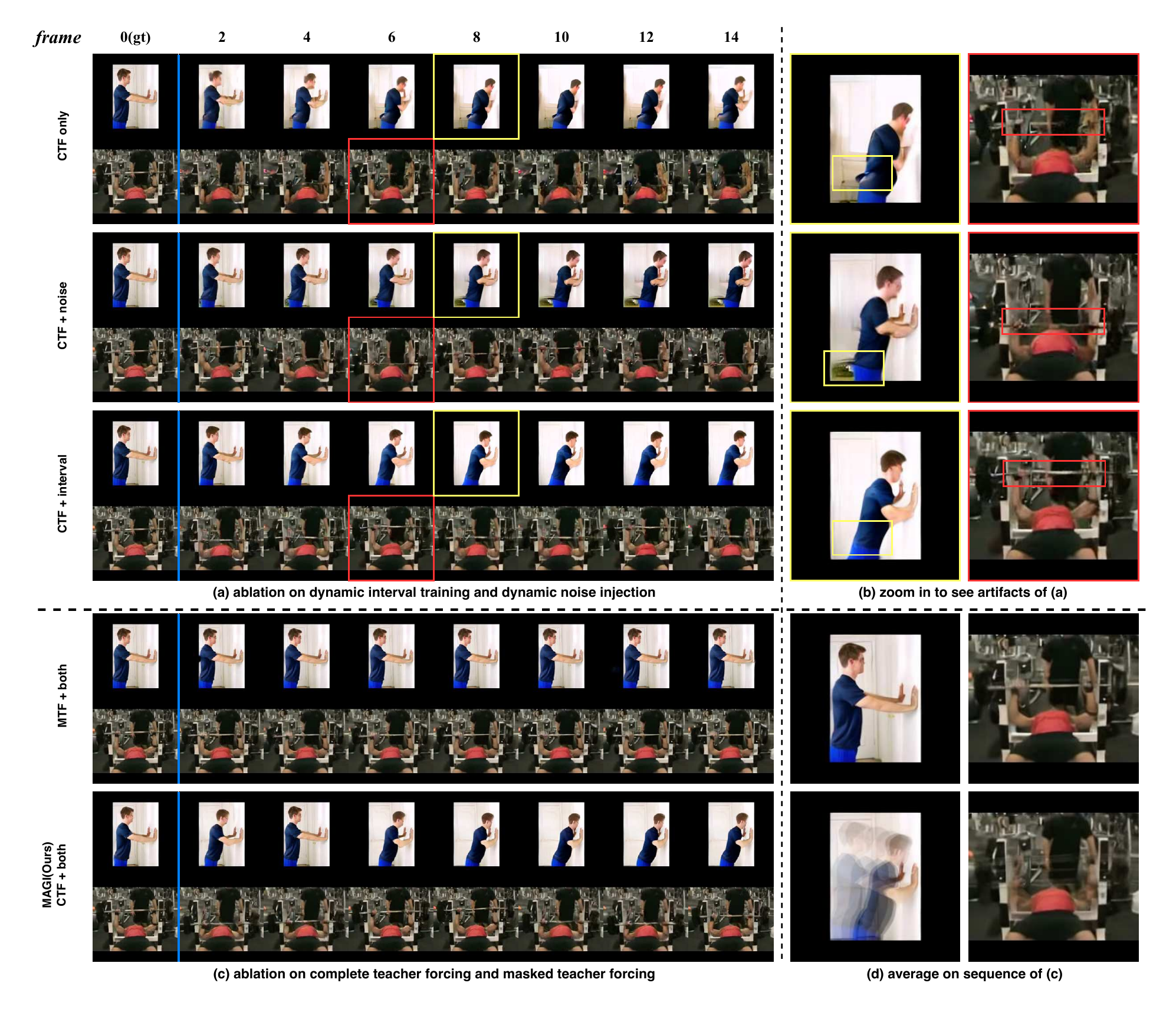}
    \vspace{-23pt}
\caption{\textbf{Case Study of Proposed Training Techniques:} This figure evaluates the impact of dynamic interval training and dynamic noise injection on CTF and MTF by: 1) visualizing CTF with and without these strategies; and 2) comparing CTF and MTF when both use them. All methods perform first-frame conditional video prediction on UCF-101~\cite{soomro2012ucf101}. The results demonstrate the efficacy of the proposed training strategies and the superior motion and temporal coherence of CTF.}
    \label{fig:ablation}
    \vspace{-2pt}
\end{figure*}
\begin{figure*}
    \centering
    \begin{subfigure}[h]{0.33\textwidth}
        \centering  
        \includegraphics[width=\textwidth]{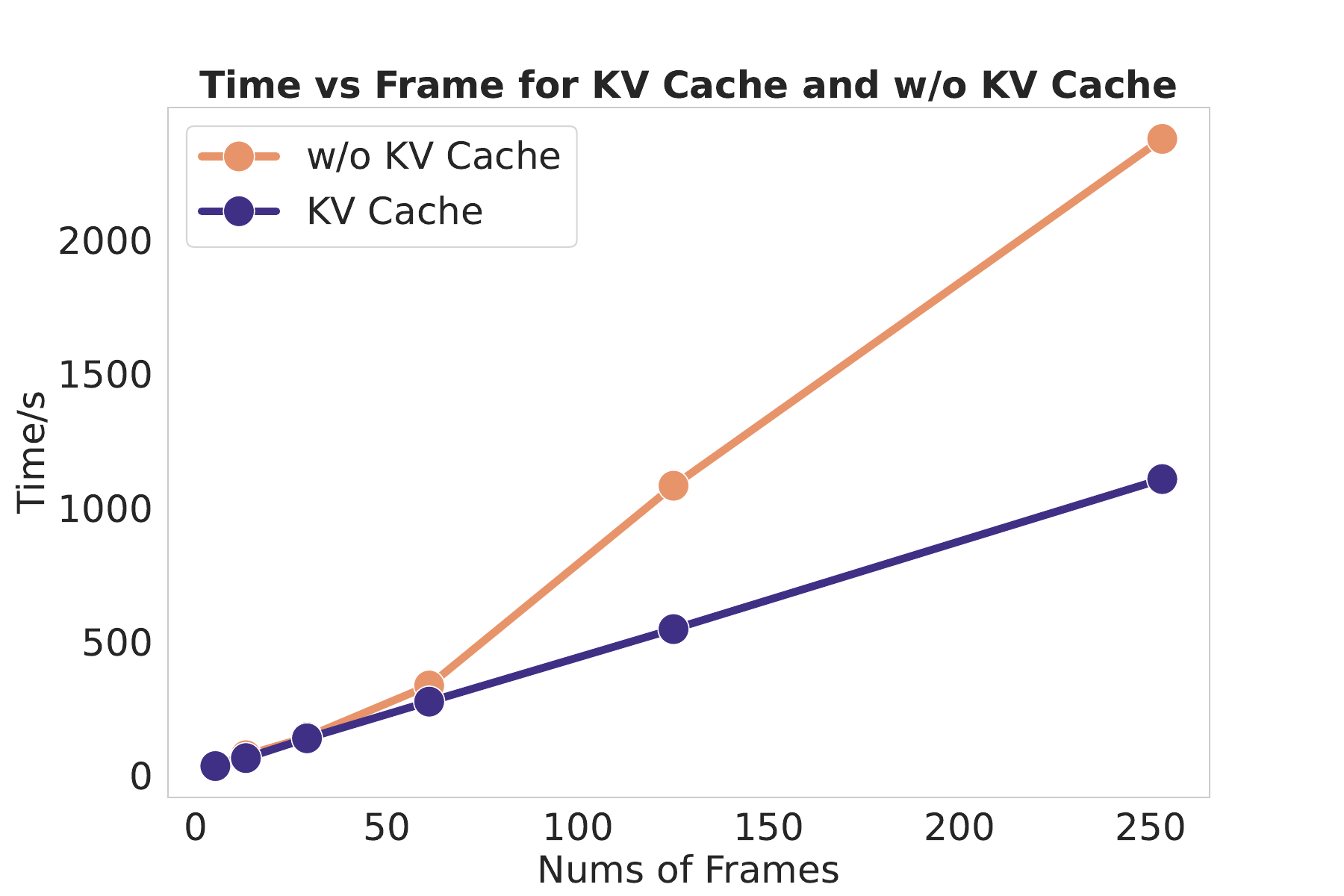} 
        \caption{Speed comparison.}
        \label{fig:kvcache_log}
    \end{subfigure}
    \begin{subfigure}[h]{0.33\textwidth}
        \centering  
        \includegraphics[width=\textwidth]{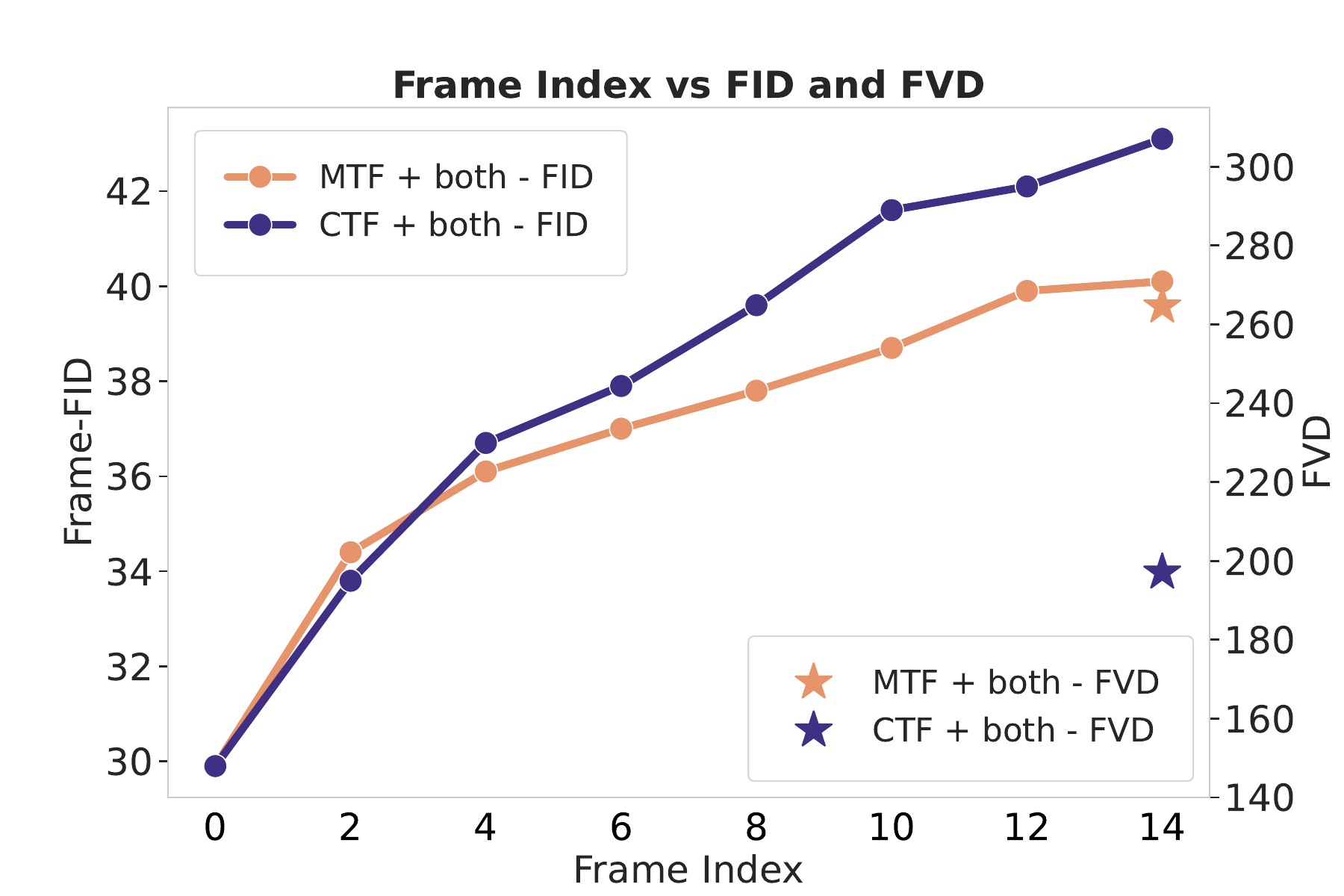}
        \caption{\textbf{MTF} ~\textbf{\textit{v.s.}} \textbf{CTF}.}
        \label{fig:ablation_teacher_forcing}
    \end{subfigure}
    \begin{subfigure}[h]{0.33\textwidth}
        \centering
        \includegraphics[width=\linewidth]{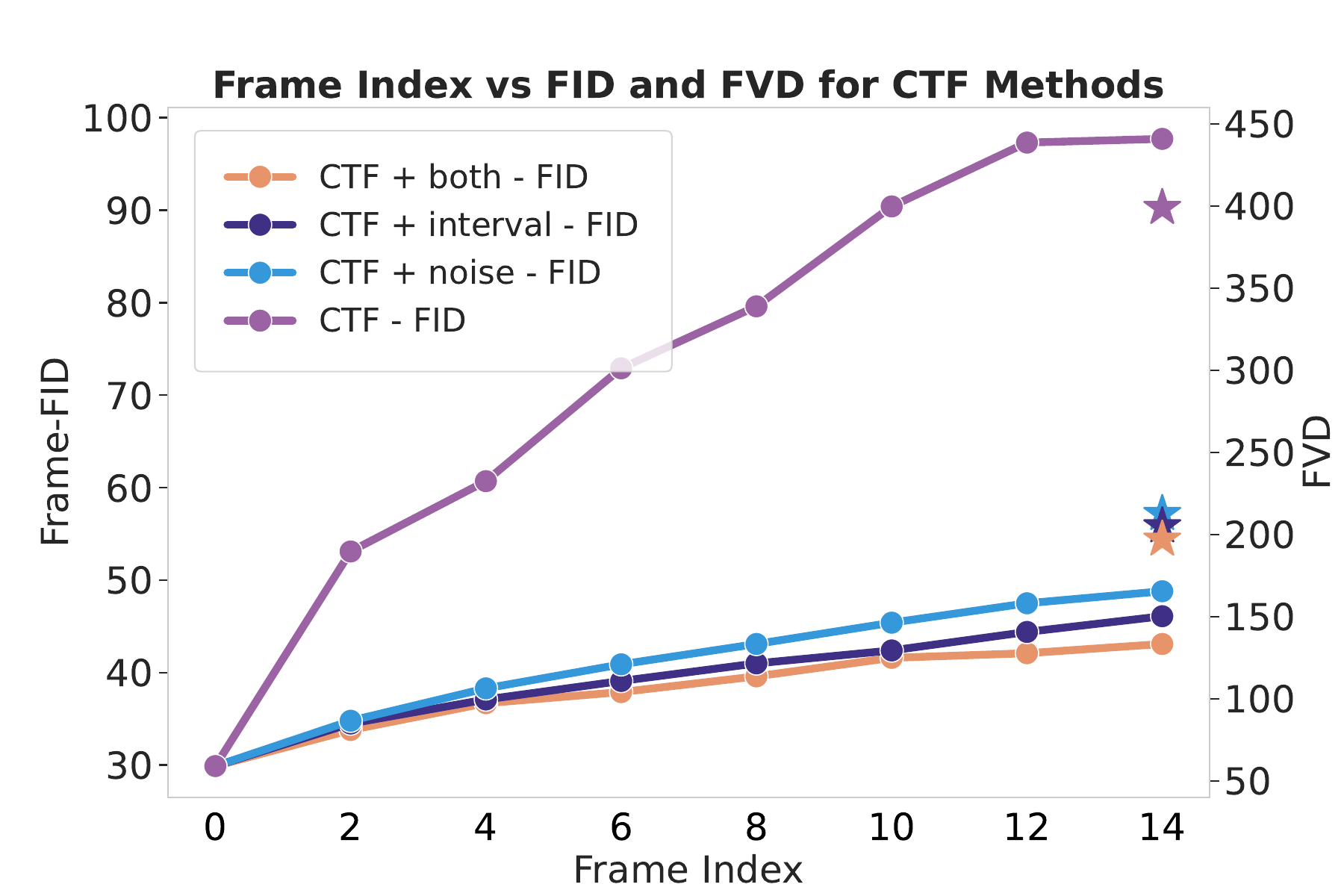}
        \caption{Ablation on first-frame conditional generation.}
        \label{fig:ablation_training_strategies}
    \end{subfigure}
    \caption{\textbf{a)} \textbf{Speed of generation process} with varying numbers of frames. \Ours achieves a significant speedup by utilizing KV Cache. \textbf{b)} \textbf{Masked Teacher Forcing (MTF)} ~\textbf{\textit{v.s.}} \textbf{Complete Teacher Forcing (CTF)}. Both methods utilize the proposed training strategies — dynamic interval training and dynamic noise injection. We report FID scores for individual frames and FVD scores for all frames on UCF101~\cite{soomro2012ucf101}. CTF achieves significantly better FVD scores but slightly worse FID scores compared to MTF. This result demonstrates that CTF better captures motion with temporal coherence, even when the quality of individual frames is lower. \textbf{c)} The results of \textbf{ablation study on first-frame conditional video predcition of UCF-101~\cite{soomro2012ucf101}}. The ``star'' indicates the FVD scores of each method with the same color.}
\end{figure*}

\section{Experiments}
\label{sec:exp}

\subsection{Experimental Setup}

\paragraph{Datasets}
For the five-frame conditional video prediction tasks, we use the Kinetics-600 dataset~\cite{carreira2018short}, which consists of 480,000 videos spanning 600 action categories. For unconditional and first-frame conditional video generation, we utilize the UCF-101 dataset~\cite{soomro2012ucf101}, containing over 13,000 clips across 101 human action classes and with totally 27 hours of recording length.

\paragraph{Implementation Details}
We set the learning rate to $2 \times 10^{-4}$ for Kinetics-600 and $1 \times 10^{-4}$ for UCF-101. Our largest model is trained for 150 epochs on Kinetics-600 and 1,400 epochs on UCF-101, with batch sizes of 256 and 128, respectively. Other hyperparameters follow the settings in MAR~\cite{li2024autoregressive}. For Kinetics-600, we use the 3D-VAE from OmniTokenizer~\cite{wang2024omnitokenizer}. For UCF-101, we use the 2D-VAE from Stable Diffusion 1.4~\cite{rombach2022high} and the 3D-VAE from Cosmos~\cite{nvidia2024cosmos}. During training, we sample 17 frames for the 3D-VAE and 16 frames for the 2D-VAE, respectively. All models are trained at a resolution of $256 \times 256$.

\paragraph{Inference and Evaluation}
For inference, we follow the strategy outlined in MAR~\cite{li2024autoregressive}, using 64 iterative steps for masked frame prediction per frame. The inference procedure is significantly accelerated using KV Cache. For evaluation on Kinetics-600, we generate 50,000 videos using context frames randomly sampled from the test set and compute the Fréchet Video Distance (FVD)~\cite{unterthiner2018towards} against the ground-truth videos, both resized to $64 \times 64$ resolution.
For UCF-101, we follow the evaluation protocol of previous works~\cite{ma2024latte,wang2024omnitokenizer}. We randomly sample 2,048 videos and compute the FVD against 2,048 randomly selected ground-truth videos from the dataset.

\subsection{Teacher Forcing Matters in Autoregressive Video Generation}
\paragraph{CTF Outperforms MTF}
Our CTF significantly outperforms MTF by nearly \textbf{23\%} on first-frame conditioned video prediction. As shown in \cref{fig:ablation_teacher_forcing}, although MTF achieves slightly better frame-wise FID, it results in substantially worse overall FVD compared to CTF. Per-frame analysis (\cref{fig:ablation}c-d) reveals that CTF better models temporal motion, while MTF generates high-quality static images lacking temporal coherence.

\paragraph{Why Does CTF Achieve Superior FVD?}
We argue that although MTF’s training on highly masked observations (using an optimal mask ratio of 70\%–100\%) is beneficial for generating visually similar frames and achieving lower FID scores, it imposes a significant limitation. Because MTF is trained to predict frames with minimal past information, it cannot effectively utilize full observations during inference. In contrast, CTF, trained on full observations, captures motion more accurately, leading to superior FVD scores.

\subsection{Addressing Exposure Bias}
\paragraph{Training Strategies Matters}
An ablation study (\cref{fig:ablation_training_strategies}) demonstrates the importance of our proposed training strategies: Removing either or both dynamic interval training and dynamic noise injection significantly degrades CTF's FVD and FID, underscoring the exposure bias in autoregressive prediction. Combining both yields the best performance, demonstrating their synergistic effect.

\paragraph{Does MTF Also Benefit from These Training Strategies?}
These training strategies prove beneficial not only for CTF but also for MTF. While we focus our quantitative analysis on CTF and omit the full results for MTF due to space limitations, similar trends of performance improvement were observed. Notably, even when MTF benefits from these strategies, CTF consistently achieves superior results (see \cref{fig:ablation_teacher_forcing}), demonstrating its inherent advantages.
\begin{table}[]
\small
\caption{\textbf{Video Prediction} on Kinetics-600~\cite{carreira2018short}. The results are evaluated on the testset on Kinetics-600. We report the $\text{FVD}_{\text{64, 50K}}$, obtained from 50K samples in the resolution of $64\times64$.
NAR: Non-autoregressive methods. AR: Autoregressive methods.}
\label{tab:kinetics_video_prediction}
\vspace{-5pt}
\centering
\resizebox{\linewidth}{!}{
\begin{tabular}{lcccc}
    \toprule[0.95pt]
    Type & Method & VAE & ${\text{FVD}}_{\text{64, 50K}}\downarrow$   \\
    \midrule[0.6pt]
    NAR & Video Diffusion~\cite{ho2022video} & - & 16.2 \\
    NAR & RIN~\cite{jabri2022scalable} & - & 10.8 \\
    NAR & MAGVIT~\cite{yu2023magvit} & MAGVIT~\cite{yu2023magvit} & 9.9 \\
    NAR & MAGVIT-v2~\cite{yu2023language} & MAGVIT-v2~\cite{yu2023language} & 4.3 \\
    AR & ViVQVAE~\cite{walker2021predicting} & VQVAE~\cite{van2017neural} & 64.3 \\
    AR & Phenaki~\cite{villegas2022phenaki} & VQVAE~\cite{van2017neural}& 36.4\\
    AR & Omni~\cite{wang2024omnitokenizer} & Omni~\cite{wang2024omnitokenizer} & 32.9 \\
    \midrule[0.6pt]
    AR & \Ours & Omni~\cite{wang2024omnitokenizer} & \textbf{11.5} \\
    \bottomrule[0.95pt]
\end{tabular}
}
\end{table}

\subsection{Benchmarking with Previous Methods}
\paragraph{Video Prediction}
We evaluate our model trained on Kinetics-600~\cite{carreira2018short} against existing non-autoregressive (NAR) and autoregressive (AR) methods (\cref{tab:kinetics_video_prediction}). 
Our method, \Ours, achieves an FVD score of \textbf{11.5}, establishing a new state-of-the-art among AR models and significantly outperforming the patch-level AR method Omni~\cite{wang2024omnitokenizer}.
Notably, our \Ours achieves -21.4 significantly lower FVD than Omni whose FVD is 32.9.
This result demonstrates the effectiveness of frame-level autoregressive modeling with Complete Teacher Forcing.
Examples of generated videos are provided in Appendix Fig.~\ref{fig:frame_pred_k600}.

\paragraph{Unconditional Video Generation}
For unconditional video generation on UCF-101~\cite{soomro2012ucf101} (\cref{tab:ucf_unconditional_video_gen}), our method achieves state-of-the-art results among AR models, outperforming Latte~\cite{ma2024latte}, a DiT-based NAR model, by approximately 50 FVD points using the same VAE. Furthermore, using a stronger VAE (Cosmos~\cite{nvidia2024cosmos}), our method becomes competitive with state-of-the-art NAR methods.

\subsection{Further Analysis}
\cref{tab:comparison_method} outlines the core differences between MAGI and other methods. 
MAGI achieves a smooth transition from patch-level to frame-level autoregressive generation while retaining key advantages, analyzed below.

\paragraph{Generation Order Matters}
Our experiments (\cref{fig:ablation_teacher_forcing} and \ref{fig:ablation_training_strategies}) demonstrate that single-frame FID increases as the number of AR steps grows, regardless of the training paradigms and strategies employed. 
This finding underscores the importance of generation order in autoregressive video generation. 
Since frame-level generation requires fewer AR steps than patch-level generation, it can better preserve image quality over long sequences by mitigating error accumulation. 
Moreover, fewer autoregressive steps reduce the propagation of errors that typically accumulate in sequential generation processes, leading to more coherent and stable video outputs. This efficiency not only enhances visual fidelity but also contributes to faster inference times, which are critical for practical applications.

\paragraph{KV Cache}
A key practical advantage of MAGI is its efficient inference. MAGI's frame-level KV Cache enables approximately linear scaling of inference time with the number of generated frames (\cref{fig:kvcache_log}), a significant advantage over the parallel computation required by NAR methods.
\begin{table}[]
\small
\caption{\textbf{Unconditional video generation} on UCF-101~\cite{soomro2012ucf101}. We report the $\text{FVD}_{\text{256, 2048}}$ which is the FVD obtained from 2048 samples in the resolution of $256\times256$.
}\label{tab:ucf_unconditional_video_gen}
\centering
\vspace{-5pt}
\resizebox{\linewidth}{!}{
\begin{tabular}{lcccc}
    \toprule[0.95pt]
    Type & Method              & VAE & $\text{FVD}_{\text{256, 2048}}\downarrow$ \\
    \midrule[0.6pt]
    NAR & MoCoGAN~\citep{tulyakov2018mocogan}    &      -      & 2886.9  \\
    NAR &MoCoGAN-HD~\citep{tian2021good} &       -     & 1729.6  \\
    NAR &DIGAN~\citep{yu2022generating}      &       -     & 1630.2  \\
    NAR &StyleGAN-V~\citep{skorokhodov2022stylegan} &      -      & 1431.0  \\
    NAR &PVDM~\citep{yu2023video}       & PVDM~\citep{yu2023video}  & 1141.9 \\
    NAR &MoStGAN-V~\citep{shen2023mostgan}  &    -        & 1380.3 \\
    NAR &Latte~\citep{ma2024latte}      & SD1.4~\cite{rombach2022high}  & 477.9 \\
    NAR &DiM~\citep{mo2024scaling} & SD1.4~\cite{rombach2022high} & 358.8 \\
    NAR &Matten~\citep{teng2024dim} & SD1.4~\cite{rombach2022high} & 210.6 \\

    AR & VideoGPT~\citep{yan2021videogpt}   & 3D VQVAE~\cite{van2017neural}  & 2880.6  \\
    \midrule[0.6pt]
    AR &\Ours & SD1.4~\cite{rombach2022high} & 420.6 & \\
    AR &\Ours & Cosmos~\cite{nvidia2024cosmos} & \textbf{297.8} & \\
    \bottomrule[0.95pt]
\end{tabular}
}
\vspace{-5pt}
\end{table}
\begin{figure*}
    \centering  
    \includegraphics[width=1.0\textwidth]{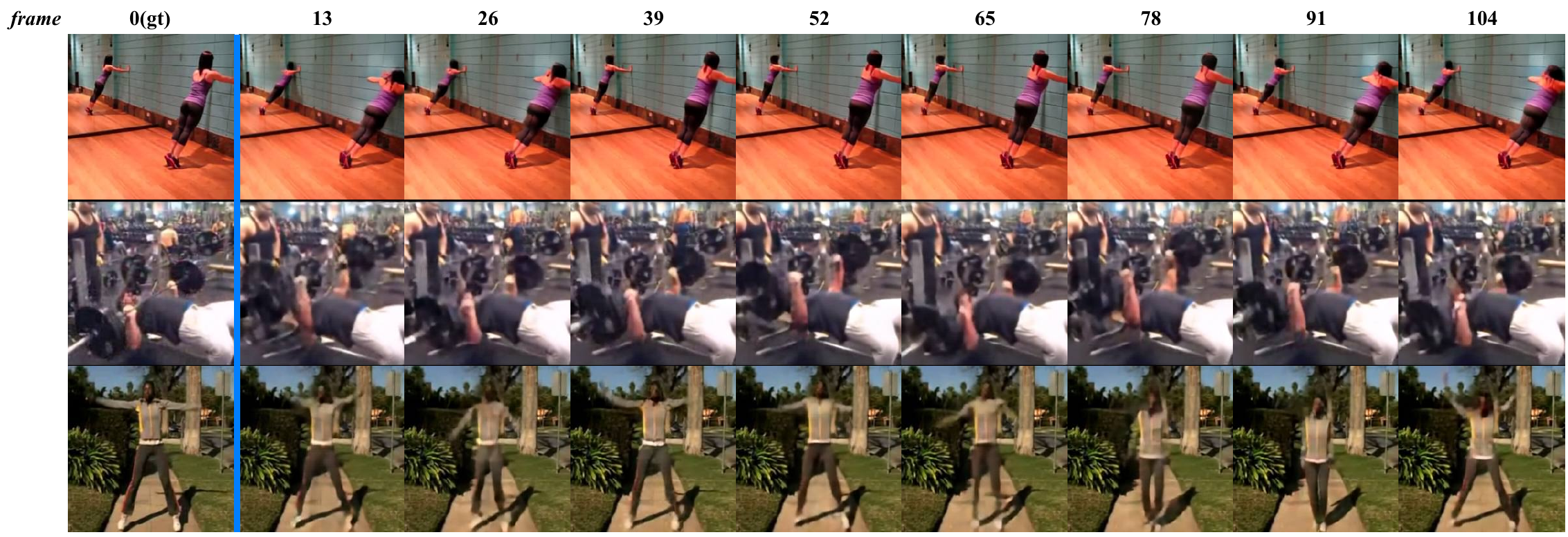}
    \vspace{-18pt}
    \caption{\textbf{Long-term Video Prediction}. \Ours predicts over 100 frames from a single input frame, maintaining reasonable motion even when trained on only 16 frames.}
    \label{fig:long_video}
    \vspace{-8pt}
\end{figure*}

\paragraph{Practical Advantages of Autoregressive Generation}
To explore this, we train MAGI on UCF-101~\cite{soomro2012ucf101} using randomly sampled 16-frame clips and evaluate its long-term prediction capability in a first-frame conditional generation setting. 
\cref{fig:long_video} demonstrates the promising predictive performance of MAGI, achieving reasonable results for sequences up to 100 frames in specific scenarios, particularly those with relatively static backgrounds and simple object motions. 
We observe performance degradation for videos with non-periodic motions (\eg, diving), which we attribute, in part, to the simplicity of the UCF-101 dataset and the challenging extrapolation required by the first-frame conditional setting. 
Specifically, for non-periodic actions like diving, the model lacks cues to predict subsequent actions once the primary action is complete, rendering metrics like FVD inappropriate. Therefore, we focus on demonstrating MAGI's long-range prediction capability in these specific scenarios to showcase its potential for capturing long-range temporal dependencies and motivate further investigation.

\section{Related Work} \label{sec:related_works}
\paragraph{Autoregressive Visual Generation}
Autoregressive language modeling~\cite{radford2018improving,radford2019language} has propelled the development of visual content generation using discrete visual tokens~\cite{van2017neural}. In this framework, pre-trained visual tokenizers like VQ-VAE~\cite{van2017neural} map visual patches into a discrete latent space, allowing visual generation to be approached similarly to language modeling. Early works such as DALL-E~\cite{DALLE} focus on text-to-image generation by learning a joint distribution between text and discrete image representations using an autoregressive cross-entropy loss. Concurrently, VideoGPT~\cite{yan2021videogpt} extends this idea to video generation, employing discrete tokens for autoregressive video prediction. VideoPoet~\cite{kondratyuk2023videopoet} further advances this approach by integrating a causal video tokenizer, MAGVIT-v2~\cite{yu2023language}, for multimodal video generation. OmniTokenizer~\cite{wang2024omnitokenizer} proposes a unified tokenizer for both discrete and continuous representations, enabling patch-level autoregressive video generation. In contrast, our work focuses on frame-level causality rather than patch-level, avoiding the limitations of raster-scan order and outperforming patch-level baselines.

\paragraph{Masked and Diffusion Video Generation}
Diffusion models have recently gained prominence in visual generation tasks~\cite{ho2020denoising,rombach2022high,he2022latent,guo2023animatediff,chen2023pixartalphafasttrainingdiffusion}, extending effectively to video generation. Video diffusion models~\cite{videoworldsimulators2024,ho2022video} employ bidirectional attention and binary mask embeddings to enable frame-level autoregressive prediction. Notable works such as GameNGen~\cite{valevski2024diffusion} use bidirectional diffusion models for real-time game generation. However, due to their bidirectional nature, these models cannot leverage KV Cache for extended video generation, limiting their scalability. Several masked video generators, such as Genie~\cite{bruce2024genie}, extend MaskGIT~\cite{chang2022maskgit} into a causal-attention-based architecture for video generation. Despite their advantages, these methods suffer from the training-inference gap inherent in masked autoregressive modeling, which negatively impacts generation quality. In contrast, our approach fully leverages KV Cache during inference, facilitated by our training paradigm that bridges the training-inference gap through a novel Complete Teacher Forcing paradigm.
\vspace{-5pt}

\paragraph{Addressing Exposure Bias for Autoregressive Video Generation}
Autoregressive models often suffer from exposure bias~\cite{zhang2019bridging} and error accumulation~\cite{laskey2017dart}, which degrade performance over long sequences. 
To mitigate these issues, we adopt noise injection from GameNGen~\cite{valevski2024diffusion}—where Gaussian noise is added to observation frame latents during training. Additionally, we introduce dynamic interval training, exposing the model to frames with varying temporal intervals to enhance generalization. As shown in \cref{fig:ablation}, both strategies individually improve robustness, and their combination yields even greater performance gains in autoregressive video generation.

\section{Conclusion}

We present MAGI, a hybrid framework that combines masked and causal modeling to achieve efficient and scalable video generation. We identify that the teacher forcing paradigm makes a significant difference—our \textbf{Complete Teacher Forcing (CTF)} approach effectively bridges the training-inference gap inherent in Masked Teacher Forcing (MTF). Additionally, we introduce essential training strategies to alleviate exposure bias. Through comprehensive experiments, we validate the effectiveness of each component. Our final model achieves state-of-the-art performance among autoregressive video generation methods. We also demonstrate the generation of long, coherent video sequences exceeding 100 frames from training sequences as short as 16 frames, highlighting MAGI’s potential as a scalable autoregressive video generation model.

\section*{Acknowledgements}
We would like to thank Tianhong Li for his insightful suggestions, invaluable help, and exceptional support. We would also like to thank Chenfei Wu, Haoyang Huang, Guoqing Ma, Hongyu Zhou, Liangyu Chen, Chunrui Han, Yimin Jiang and Yu Deng for their constructive discussions and advice, which greatly improved this work.

{
    \small
    \bibliographystyle{ieeenat_fullname}
    \bibliography{main}
}

\newpage
\appendix
\section{Additional Details}
\label{sec:additional_details}
\begin{figure*}[h!]
    \centering  
    \includegraphics[width=1.0\textwidth]{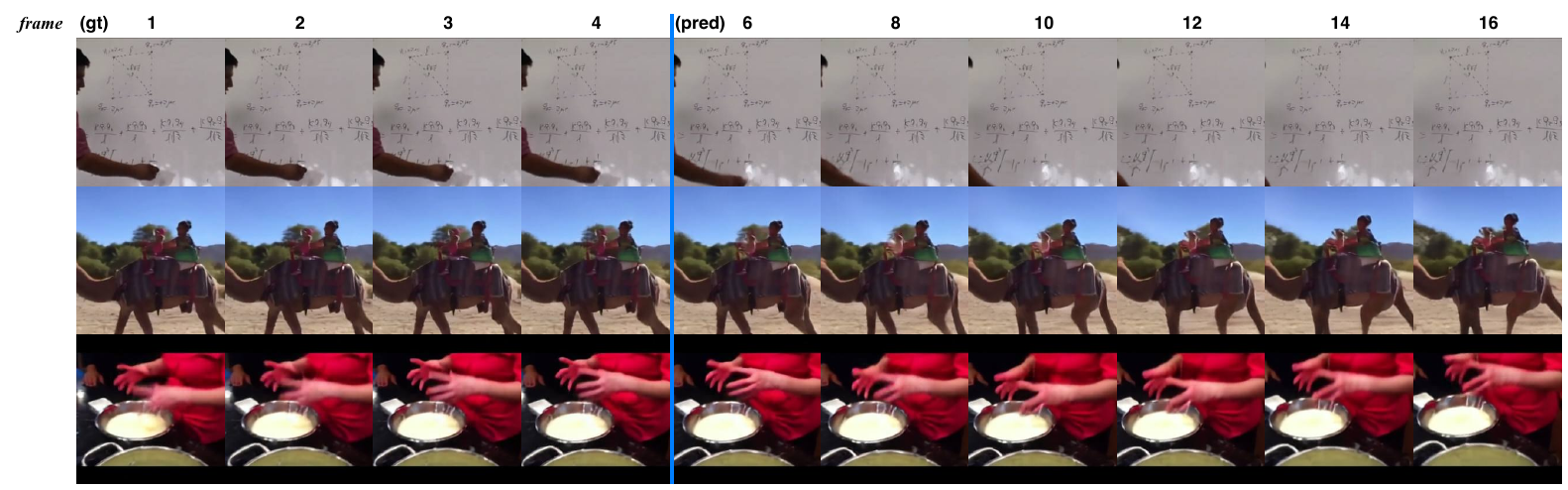}
    \caption{\textbf{Case Study: Video Prediction on Kinetics-600}. MAGI generates high-quality future frames conditioned on past frames.}
    \label{fig:frame_pred_k600}
\end{figure*}

\subsection{Network Architecture}
\label{sec:network_architecture}

Our spatial-temporal transformer comprises 20 spatial-temporal blocks. Each spatial-temporal block consists of a spatial layer followed by a temporal layer. The specific configurations are as follows:
\begin{itemize}
    \item \textbf{Hidden States Dimension:} 1280
    \item \textbf{MLP Dimension in Attention Layers:} 5120
    \item \textbf{Number of Attention Heads:} 16
    \item \textbf{Diffusion Head:} Comprises 3 blocks with a dimension of 1024, implemented identically to the diffusion head in MAR~\cite{li2024autoregressive}.
\end{itemize}

Overall, our model consists of approximately \textbf{850} million parameters, which is on the same scale as the baseline models, ensuring a fair comparison.

\subsection{Data Sampling}
\label{sec:data_sampling}

\subsubsection{Training}
During training, we uniformly sample a video from the training set and subsequently extract a clip based on a randomly selected frame interval. Specifically, with dynamic interval training, the frame interval is uniformly sampled from 1 to 25. 

\subsubsection{FVD Evaluation}
We use different sampling strategies for the UCF-101 and Kinetics-600 (K600) datasets for the real distribution:

\begin{itemize}
    \item \textbf{UCF-101:} We randomly sample 2,048 videos and extract a single clip from each using a fixed frame interval of 3.
    \item \textbf{K600:} We randomly sample 50,000 videos and extract a single clip from each with a frame interval of 1.
\end{itemize}

\noindent We also generate a equal number of clips for FVD computation. It is important to note that MAGVIT-1~\cite{yu2023magvit} utilizes a real distribution of 300,000 videos by applying 6 random spatial and temporal crops per video in the evaluation on K600. To maintain simplicity and ensure reproducibility, we limit our evaluation to 50,000 videos for the real distribution. Consequently, our Fréchet Video Distance (FVD) score does not benefit from the larger sample size used in MAGVIT-1.

\subsection{Training Procedure}
\label{sec:training_procedure}

The training process involves the following configurations:
\begin{itemize}
    \item \textbf{Warmup Steps:} 10,000
    \item \textbf{Learning Rate:} $2 \times 10^{-4}$
    \item \textbf{Weight Decay:} 0.02
    \item \textbf{Input Frames:}
    \begin{itemize}
        \item 17 frames when using a 3D-VAE
        \item 16 frames when using a 2D-VAE
    \end{itemize}
    \item \textbf{Batch Size:}
        256 for K600 and 128 for UCF-101. 
    \item \textbf{Training Epoch:}
        150 for K600 and 1400 for UCF-101. 
    \item \textbf{Patch Size in Input Layer:} 2 (consistent across all experiments)
    \item \textbf{Exponential Moving Average (EMA):} Applied with a decay rate of 0.9999. All reported results are generated using the EMA model.
    \item \textbf{Training Precision:} BF16, which is crucial for model convergence.
\end{itemize}

\subsection{Inference Procedure}
\label{sec:inference_procedure}

During inference, the following settings are applied:
\begin{itemize}
    \item \textbf{Default Masked Prediction Steps:} 64
    \item \textbf{Denoising Steps for Diffusion Head:} 100
\end{itemize}

It is important to note that:
\begin{itemize}
    \item During training, each spatial layer employs fully bidirectional attention without any attention mask, while each temporal layer utilizes causal attention with a designed attention mask.
    \item At inference time, the attention mask is unnecessary as the model attends only to previously generated frames.
    \item When inferring longer sequences, we interpolate the temporal position embeddings using linear interpolation.
    \item In our proposed CTF paradigm, the sequence length of the model input is doubled solely during training. The sequence length remains unchanged during inference.
\end{itemize}

\section{Future Work}
\label{sec:future_work}

Future research directions include:
\begin{itemize}
    \item \textbf{Text-to-Video Generation:} Extend our approach to text-to-video tasks and explore the scalability of both model and data sizes.
    \item \textbf{Interactive Game Generation:} Apply our model to real-time interactive game generation within dynamic environments, presenting a promising new direction.
\end{itemize}
\end{document}